\documentclass[letterpaper, 10 pt, conference]{ieeeconf}  % Comment this line out if you need a4paper

\IEEEoverridecommandlockouts                              % This command is only needed if 
\usepackage{times}

\usepackage[bookmarks=true]{hyperref}
\usepackage{multicol}
\usepackage{booktabs}
\usepackage[weather]{ifsym}
\usepackage{siunitx}
\usepackage{tabularx}

\usepackage[dvipsnames]{xcolor}
\usepackage{moreverb,url}
\usepackage{amsmath,amsfonts,amssymb}
\usepackage{tikz,tkz-euclide}
\usepackage{balance}
\usetikzlibrary{decorations,calligraphy}
\usetikzlibrary{calc}
\tikzset{>=latex}
\usepackage{multicol}
\usepackage{multirow}
\usepackage{tabularx}
\usepackage{booktabs}
\usepackage{siunitx}
\newcolumntype{Y}{>{\centering\arraybackslash}X}

\usepackage[nolist,nohyperlinks]{acronym}

\def\state{{\mathcal{S}}}

\newcommand\trans[2]{{\mathbf{T}_{#1}^{#2}}}
\newcommand\transhat[2]{{\hat{\mathbf{T}}_{#1}^{#2}}}
\newcommand\transtilde[2]{{\tilde{\mathbf{T}}_{#1}^{#2}}}
\def\bias{{b}}
\def\biastilde{{\tilde{b}}}

\newcommand\pos[2]{{\mathbf{p}_{#1}^{#2}}}
\newcommand\poshat[2]{{\hat{\mathbf{p}}_{#1}^{#2}}}

\def\state{{\mathcal{S}}}
\def\statehat{{\hat{\mathcal{S}}}}
\def\world{{W}}
\def\gt{{\text{GT}}}

\newcommand\resbias[1]{{r_{b_{#1}}}}
\newcommand\infobias[1]{{\Omega_{b_{#1}}}}

\newcommand\resrel[2]{{\mathbf{r}_{#1}^{#2}}}
\newcommand\inforel[2]{{\boldsymbol{\Omega}_{#1}^{#2}}}

\def\coordinate{{\mathbf{e}}}

\def\bilinear{\beta}
\def\corr{g}

\def\localmap{\mathcal{M}}

\def\inv{^{\text{-}1}}
\def\loops{\mathcal{L}}
\def\Log{\text{Log}}

\begin{acronym}[AAAAAAAAA]
    \acro{ate}[ATE]{Absolute Trajectory Error}
    \acro{drpogo}[Dr-PoGO]{Direct Radar Pose-Graph Optimization}
    \acro{epe}[EPE]{End-Pose Error}
    \acro{fmcw}[FMCW]{Frequency-Modulated Continuous-Wave}
    \acro{rmse}[RMSE]{Root Mean Squared Error}
    \acro{slam}[SLAM]{Simultaneous Localization And Mapping}
    \acro{SE2}[SE(2)]{Special Euclidean group in two dimensions}
\end{acronym}

\title{\LARGE \bf
Dr-PoGO: Direct Radar Pose-Graph Optimization
}

\author{Cedric Le Gentil$^{1}$, Weican Li$^{1}$, Leonardo Brizi$^{2}$, Timothy D. Barfoot$^{1}$% <-this % stops a space
\thanks{$^{1}$~Robotics Institute University of Toronto.
        {Corresponding author: \tt\small cedric.legentil@robotics.utias.utoronto.ca}}%
\thanks{$^{2}$~Department of Computer, Control, and Management Engineering “Antonio Ruberti”, Sapienza University of Rome.}
\thanks{© 2026 IEEE.  Personal use of this material is permitted.  Permission from IEEE must be obtained for all other uses, in any current or future media, including reprinting/republishing this material for advertising or promotional purposes, creating new collective works, for resale or redistribution to servers or lists, or reuse of any copyrighted component of this work in other works.}
}

\begin{document}

\maketitle
\thispagestyle{empty}
\pagestyle{empty}

%%%%%%%%%%%%%%%%%%%%%%%%%%%%%%%%%%%%%%%%%%%%%%%%%%%%%%%%%%%%%%%%%%%%%%%%%%%%%%%%
\begin{abstract}

This paper introduces Dr-PoGO, a method for Simultaneous Localization And Mapping (SLAM) using a 2D spinning radar.
Unlike cameras or lidars that require line-of-sight, millimetre-wave radars can `see' through dust, falling snow, rain, etc.
Accordingly, it is a great modality for robust perception regardless of the weather conditions.
While most existing radar-based SLAM methods rely on the extraction of point clouds or features to perform ego-motion estimation, Dr-PoGO leverages direct registration techniques for odometry (DRO) and loop-closure registration.
An off-the-shelf radar-focused place recognition algorithm, RaPlace, provides loop-closure candidates.
As RaPlace does not provide relative transformations, Dr-PoGO introduces a coarse-to-fine registration that uses visual features and descriptors to obtain an initial guess for the direct transformation refinement.
The global trajectory is optimized in a pose-graph optimization.
Dr-PoGO demonstrates state-of-the-art performance over \SI{300}{\kilo\meter} of data in various real-world automotive environments.
Our implementation is publicly available: \url{https://github.com/utiasASRL/dr_pogo}.

\end{abstract}

%%%%%%%%%%%%%%%%%%%%%%%%%%%%%%%%%%%%%%%%%%%%%%%%%%%%%%%%%%%%%%%%%%%%%%%%%%%%%%%%
\section{Introduction}

At the forefront of any autonomous stack, robust perception and state estimation are required to guarantee reliable robotic operations regardless of the environment and weather conditions.
Radar-based sensing was the core of early landmark-based \ac{slam} works~\cite{dissanayake2001slam}.
With the popularization of millimetre-wave radars in the automotive industry for advanced driver assistance systems, there has been a renewed interest in this modality among the robotics community~\cite{harlow2024newwave}.
While radar data are generally less accurate than lidar, they can `see' through dust, smoke, heavy precipitation, etc., providing the required robustness for all-weather navigation~\cite{legentil2025dowe}. 
This paper introduces \ac{drpogo}, a method for 2D trajectory estimation using \ac{fmcw} radars.

\begin{figure}
    \def\legenddist{0.1cm}
    \def\dataheight{3.0cm}
    \def\vdist{4.2cm}
    \centering
    \begin{tikzpicture}
        \tikzstyle{legend} = [align = center, inner sep=0, outer sep=0, node distance = 0em, execute at begin node=\setlength{\baselineskip}{8pt}\small] 
        \node[inner sep = 0, outer sep = 0] (boreas) {\includegraphics[clip,width = \columnwidth]{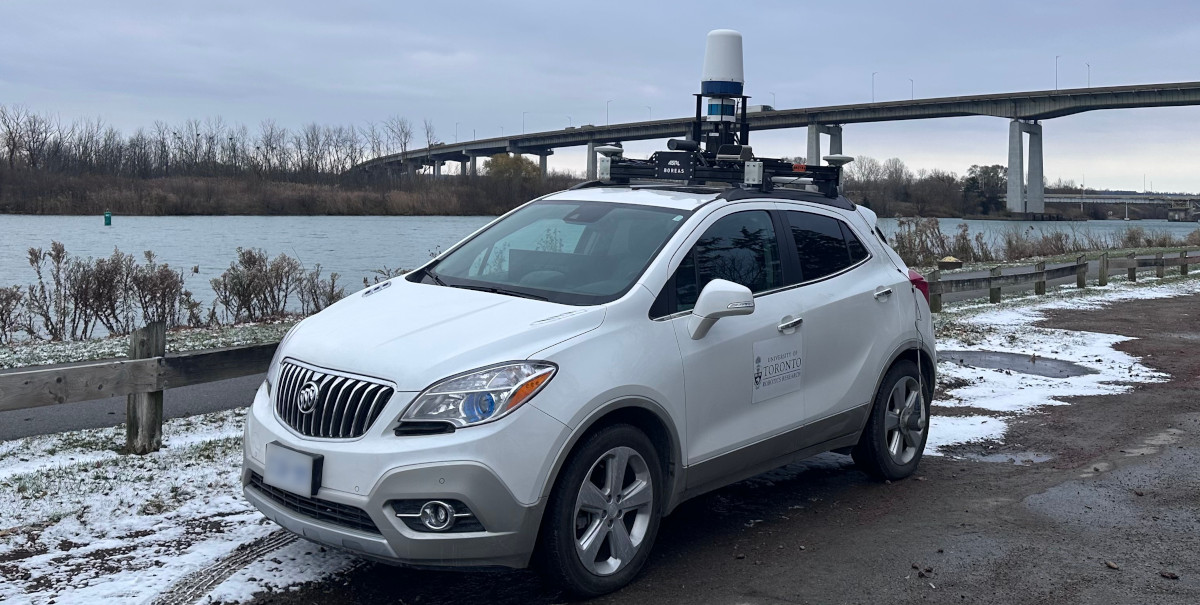}};
        \node[legend, below=\legenddist of boreas] {(a) Radar-inertial-lidar-visual automotive collection platform};
        \node[inner sep=0, outer sep=0, below=\vdist of boreas.west, anchor=west] (polar){\includegraphics[clip, height = \dataheight]{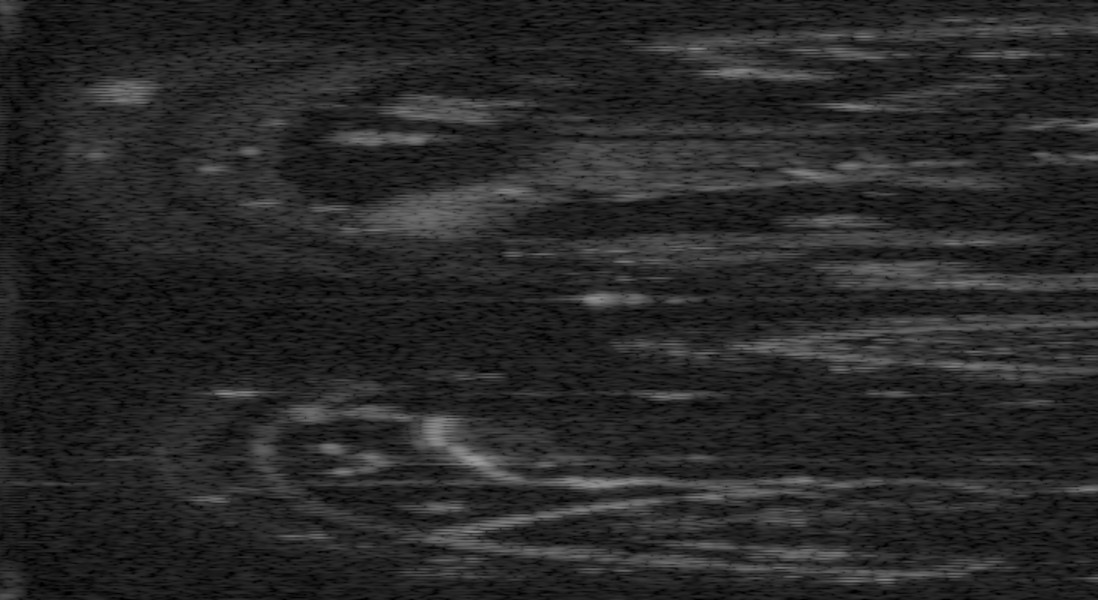}};
        \node[inner sep=0, outer sep=0, below=\vdist of boreas.east, anchor=east] (cart){\includegraphics[clip, height = \dataheight]{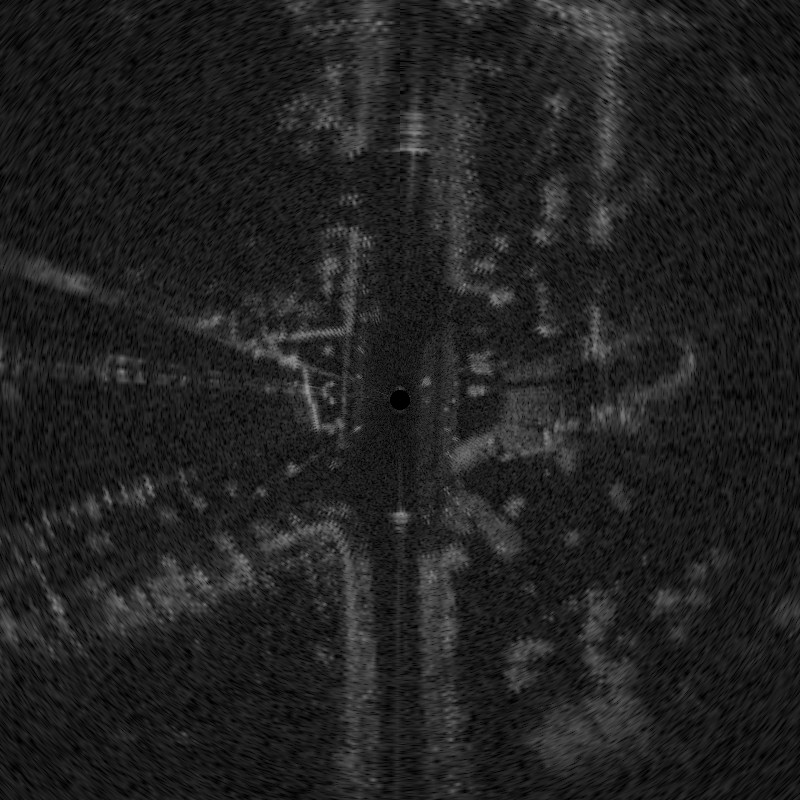}};
        \node[legend, below=\legenddist of polar] {(b) Raw radar polar image};
        \node[legend, below=\legenddist of cart] {(c) Cartesian image};
    \end{tikzpicture}
    %\vspace{-0.3cm}
    \caption{Dr-PoGO performs SLAM using 2D radar data. It is validated with \SI{225}{\kilo\meter} of real-world data. (a) is the platform used to record data (in front of the \texttt{Skyway} environment). (b) and (c) are a sample of radar data in polar and Cartesian forms, respectively.}
    \label{fig:teaser}
\end{figure}

Radar-based state estimation is commonly performed by first extracting point clouds or features from the raw data before leveraging ICP-like registration techniques.
While lightening the computational burden, the feature extraction process disregards a non-negligible amount of information originally contained in the raw data.
Direct Radar Odometry (DRO)~\cite{legentil2025dro} demonstrated superior accuracy by performing direct scan-to-local-map registration through continuous cross-correlation using all the scan's information.
In this work, we leverage DRO for odometry before full trajectory optimization, and we extend the use of direct methods to local-map-to-local-map registration.

Whether it is exteroceptive or proprioceptive-based, or both, odometry is bound to drift due to sensor noise, outlier associations, numerical precision, etc.
Pose-graph optimization provides a way to correct for the accumulated drift by explicitly integrating \emph{loop closures} in a global optimization that estimates the whole system's trajectory, not only the current pose.
Accordingly, loop-closure detection, detecting when the robot revisits previously observed areas, has been a strong focus to improve state estimation accuracy.
Radar-based perception is no exception, and multiple works have addressed this problem.
\ac{drpogo} leverages RaPlace~\cite{jang2023raplace} to detect loop closures by matching together pairs of radar local-maps built with DRO.
Unfortunately, RaPlace does not provide the metric relative transformation between matching local maps.
To address this gap, we extend DRO's abilities to perform local-map-to-local-map registration, as aforementioned.
However, unlike in the odometry scenario, where consecutive scans are collected almost from the same location and thus where the identity transformation is a sufficient initial guess, potential loop-closure pairs can correspond to quite different robot poses.
\ac{drpogo} extracts keypoints and their associated descriptors to perform RANSAC-based coarse registration before a refinement with a direct approach.
The use of descriptor-based associations alleviates the need for an initial guess.

This work (i) introduces a novel pipeline mixing feature-based and direct methods for radar data registration without initial guess; (ii) integrates direct odometry, loop-closure detection, and registration in a radar \ac{slam} formulation built upon pose-graph optimization; (iii) demonstrates state-of-the-art performance over more than \SI{300}{\kilo\meter} of public and self-collected real-world data. Our platform and a data sample are shown in Fig.\ref{fig:teaser}.

\section{Related work}

Radars used for robotics perception can generally be split into two main categories: phased arrays for 3D/4D data collection and mechanically spinning antennas for \SI{360}{\degree} 2D sensing. 
As \ac{drpogo} relies on the latter, this section focuses on 2D radar methods and state estimation over the \ac{SE2}.
For more details on 3D/4D radar perception, we invite the reader to refer to~\cite{harlow2024newwave}, \cite{nader2024survey}, and~\cite{venon2022millimeter}.
The data collected with \ac{fmcw} 2D spinning radars consist of polar images where each row corresponds to an azimuth and each column to a different range.
As shown in Fig.~\ref{fig:teaser}, the data can be converted into \emph{Cartesian images} for visualization or processing.
The intensity of the pixels corresponds to the reflections of the radar-emitted radio waves on elements of the environment.

Modern radar-based methods have mostly relied on the extraction of point clouds or features from the raw data.
Early works used standard computer-vision keypoints and descriptors~\cite{lowe2004sift} over Cartesian images~\cite{callmer2011radarslamusingvisualfeatures}.
Feature extractors such as CFAR~\cite{rohling1983radarcfar}, K-strongest~\cite{adolfsson2021CFEAR}, or CFEAR~\cite{adolfsson2023cfear} are used to extract points for each azimuth from the polar images.
The authors of~\cite{preston2025finer} evaluated the impact of feature extraction on radar-based odometry~\cite{preston2025finer}.
It is important to note that radar data is not collected as a snapshot.
As the radar's antenna sweeps its radio-wave beam around (generally at around \SI{4}{\hertz}), any motion of the robot will create motion distortion in the collected scans.
Atop the simple fact that the sensor is physically at two different locations when collecting two consecutive azimuths, the relative velocity between the sensor and the environment leads to further distortion due to the Doppler effect.
These sources of distortion can be accounted for in a `one-off' undistortion step based on previous state estimates, as in~\cite{adolfsson2023cfear}.
Another approach is to use continuous-time state estimation to elegantly undistort radar-extracted points at the same time as estimating the platform's ego-motion, such as~\cite{burnett2021dowe}, \cite{are_we_ready_for}, and~\cite{burnett2025continuous}).
Other methods leverage deep-learned features~\cite{hero_paper} or weights~\cite{Lisus2025Pointing} to robustify the scan registration process.

Another branch of radar-based works uses \emph{direct} methods for scan registration.
The most common approach is to compute cross-correlation scores between the current scan and the previous one for a discrete set of potential state updates
The hypothesis with the largest score is used to update the state.
Works such as~\cite{Checchin2009}, \cite{masking_by_moving}, and \cite{park2020pharao} rely on this principle.
Note that these methods do not model motion distortion and the Doppler effect.
The authors of~\cite{lisus2025doppler} used a radar with a triangular modulation pattern and demonstrated that estimating the radial velocities from the observed Doppler distortion provided a reliable source of odometry, even in feature-deprived environments, when associated with a yaw-gyroscope.
Recently, \cite{legentil2025dro} introduced DRO, the first continuous direct method that rigorously accounts for motion and Doppler distortions by maximizing a continuous cross-correlation score between the current scan and a local map that is updated `on-the-fly'.

Full-batch trajectory optimization with loop-closure constraints is a way to mitigate the inherent drift of odometry.
In its simplest form, the trajectory is estimated by optimizing a pose graph where each link is a constraint on the relative pose of the sensor at two different timestamps.
Pose-graph optimization is the backbone of radar \ac{slam} works such as~\cite{holder2019realtime}, \cite{hong2022radarslam}, and \cite{adolfsson2023tbv}.
These rely on loop-closure detection approaches originally designed for lidar data (\cite{himstedt2014largescale}, \cite{he2016m2dp}, and~\cite{kim2022scancontextpp}, respectively).
\ac{drpogo} also builds upon a pose graph for full-batch trajectory estimation, but it leverages a radar-specific place-recognition algorithm: RaPlace~\cite{jang2023raplace}.
RaPlace employs the Radon Transform over Cartesian radar scans and Fast Fourier Transforms to compute rotation- and translation-invariant scan descriptors.
Loop candidates are selected by thresholding the cross-correlation between descriptors.
To the best of our knowledge, \ac{drpogo} is the first radar-based \ac{slam} framework built using direct methods for relative pose estimation, both for odometry and loop-closure constraints.

\section{Method}

\subsection{Overview}

Let us consider a 2D spinning \ac{fmcw} radar rigidly mounted to a ground robotic platform.
The radar's spinning axis is orthogonal to the locally planar motion of the platform.
Optionally, a yaw gyroscope is also rigidly mounted to the platform with its axis parallel to the radar's spinning axis.
The proposed method, \ac{drpogo}, estimates the trajectory of the radar $\trans{\world}{t_m} \in \mathrm{SE}(2)$, with $m = 1,\cdots, N$ and $N$ the total number of radar scans in a given sequence.

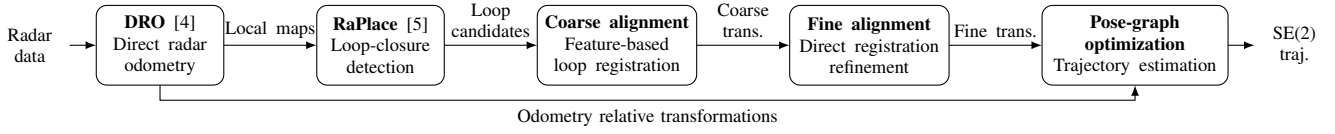
\begin{figure*}
    \centering
    \begin{tikzpicture}
        \def\hdist{5em}
        \def\vdist{3.5em}
        \def\vdistshort{1.0em}
        \def\blockheight{3.0em}
        \def\blockwidth{6.0em}
        \def\innerpad{0.0em}
        \tikzstyle{io} = [text width = 3em,  minimum width = 3em, align = center, inner sep=\innerpad, outer sep=0, node distance = 0em, execute at begin node=\setlength{\baselineskip}{8pt}\scriptsize] 
        \tikzstyle{block} = [draw, fill=white, rectangle, minimum height = \blockheight, text width = \blockwidth,  minimum width = \blockwidth, align = center, inner sep=\innerpad, outer sep=0, node distance = 0em, execute at begin node=\setlength{\baselineskip}{8pt}\scriptsize, rounded corners]
        \tikzstyle{shorterblock} = [draw, fill=white, rectangle, minimum height = \blockheight, text width = 0.8*\blockwidth,  minimum width = 0.8*\blockwidth, align = center, inner sep=\innerpad, outer sep=0, node distance = 0em, execute at begin node=\setlength{\baselineskip}{8pt}\scriptsize, rounded corners]
        \tikzstyle{longerblock} = [draw, fill=white, rectangle, minimum height = \blockheight, text width = 7.0em,  minimum width = 7.0em, align = center, inner sep=\innerpad, outer sep=0, node distance = 0em, execute at begin node=\setlength{\baselineskip}{8pt}\scriptsize, rounded corners] 
        \tikzstyle{arrowlabel} = [text width = \vdist,  minimum width = \vdist, align = center, inner sep=\innerpad, outer sep=0, execute at begin node=\setlength{\baselineskip}{8pt}\scriptsize, above=1mm]

        \node[shorterblock] (dro) {\textbf{DRO} \cite{legentil2025dro} \\ Direct radar odometry};
        \node[shorterblock, right=\vdist of dro] (raplace) {\textbf{RaPlace} \cite{jang2023raplace} \\ Loop-closure detection};
        \node[block, right=\vdist of raplace] (coarse) {\textbf{Coarse alignment} \\ Feature-based loop registration};
        \node[block, right=\vdist of coarse] (fine) {\textbf{Fine alignment} \\ Direct registration refinement};
        \node[longerblock, right=\vdist of fine] (pogo) {\textbf{Pose-graph optimization} \\ Trajectory estimation};

        \node[io, left=\vdistshort of dro] (input) {Radar data};
        \node[io, right=\vdistshort of pogo] (output) {SE(2) traj.};

        \draw[->] (input) -- (dro);
        \draw[->] (dro) -- node[arrowlabel]{Local maps} (raplace);
        \draw[->] (raplace) -- node[arrowlabel]{Loop candidates} (coarse);
        \draw[->] (coarse) -- node[arrowlabel]{Coarse trans.} (fine);
        \draw[->] (fine) -- node[arrowlabel]{Fine trans.} (pogo);
        \draw[->] (pogo) -- (output);
        \coordinate[below=2mm of dro] (waypoint1);
        \coordinate[below=2mm of pogo] (waypoint2);
        \draw[->] (dro.south) -- (waypoint1) -- node[below]{\scriptsize Odometry relative transformations} (waypoint2) -- (pogo.south);
    \end{tikzpicture}
    \vspace{-0.4cm}
    \caption{Block-diagram overview of Dr-PoGO. First, DRO processes the raw radar data (and optional yaw-gyroscope measurements) to generate per-scan local maps and relative SE(2) transformations. Leveraging the local maps, RaPlace proposes loop-closure candidates that are later used for coarse-to-fine registration. Finally, a pose-graph optimization estimates the whole trajectory with DRO's relative poses and the refined loop-closure transformations.}
    \label{fig:overview_diagram}
\end{figure*}

Fig.~\ref{fig:overview_diagram} shows the different blocks of \ac{drpogo}.
First, direct radar odometry is performed with DRO~\cite{legentil2025dro}.
The latter estimates the system's ego-motion $\transtilde{t_m}{t_{m+1}}$ through radar-based scan-to-local-map registration and updates the local map for each registered scan.
\ac{drpogo} leverages both DRO's scan-to-scan motion estimates $\transtilde{t_m}{t_{m+1}}$ and local maps $\localmap_m$.
Loop closures are detected using RaPlace~\cite{jang2023raplace} between local maps in Cartesian form.
Succinctly, RaPlace computes the Radon transform of each incoming scan (or local map) before evaluating its cross-correlation with past scans (or local maps).
It results in a set of matching scores, one per past scan (or local map).
\ac{drpogo} selects the past local map with the highest score to form a pair $\{\localmap_i, \localmap_j\}$ that represents a potential loop closure without thresholding the matching score.
The goal is to maximize the number of true-positive matches that are carried to the next step.
For each pair $\{\localmap_i, \localmap_j\}$, \ac{drpogo} attempts local-map-to-local-map registration in a coarse-to-fine manner as detailed in Section~\ref{sec:closure_registration}.
Finally, successful loop-closure transformations $\transtilde{t_i}{t_j}$ are integrated in a pose-graph batch optimization problem that estimates the full trajectory $\statehat = \{\transhat{\world}{t_1},\cdots,\transhat{\world}{t_N}\}$.
The motivation to perform \ac{drpogo} using DRO's on-the-fly local maps instead of radar scans is their robustness to noise and to the presence of dynamic objects in the environment.
Built using per-pixel low-pass filters, the local maps reliably capture the static elements of the scene, thus enabling high-quality loop-closure detection and registration in real-world driving scenarios where inconsistent traffic can hinder scan-based approaches. 

\subsection{Loop-closure direct registration}
\label{sec:closure_registration}

DRO~\cite{legentil2025dro} demonstrated that direct radar data registration outperformed feature-based methods in terms of registration accuracy.
However, as DRO's cross-correlation maximization is not performed with a global solver, it requires a decent initial guess to converge toward the true transformation between two local maps.
While it is easily obtained for consecutive scan registration (odometry), this is not the case for loop-closure registration due to the accumulated drift along the sensor's trajectory.
Accordingly, \ac{drpogo} first uses a coarse feature-based registration step before refining the relative pose estimate with a direct approach similar to DRO's.

\subsubsection{Feature-based coarse registration}

Given two local maps $\localmap_i$ and $\localmap_j$ as Cartesian intensity images, SIFT features and descriptors~\cite{lowe2004sift} are extracted independently in both images.
Using a brute-force matcher between the two descriptor sets, a list of candidate matches is obtained.
The coarse transformation between $\localmap_i$ and $\localmap_j$ is estimate through RANSAC-based~\cite{fischler1981ransac} \ac{SE2} registration.
Note that no motion-distortion and Doppler-shift corrections are needed, as these are already accounted for during the local-map generation in DRO.
The number of inliers and a user-defined threshold are used to reject unlikely loop closures.

\subsubsection{Direct fine registration}

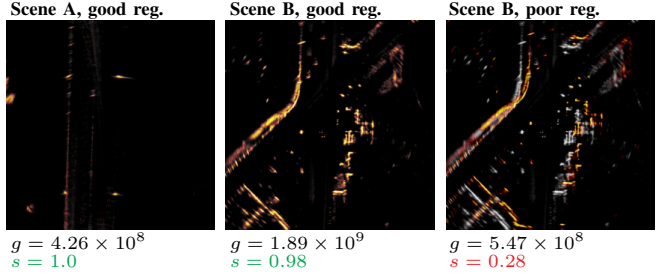
\begin{figure}
    \centering
    \begin{tikzpicture}
        \node[inner sep=0, outer sep=0]{\includegraphics[clip, width = \columnwidth, trim=0.3cm 1cm 0.3cm 1.0cm]{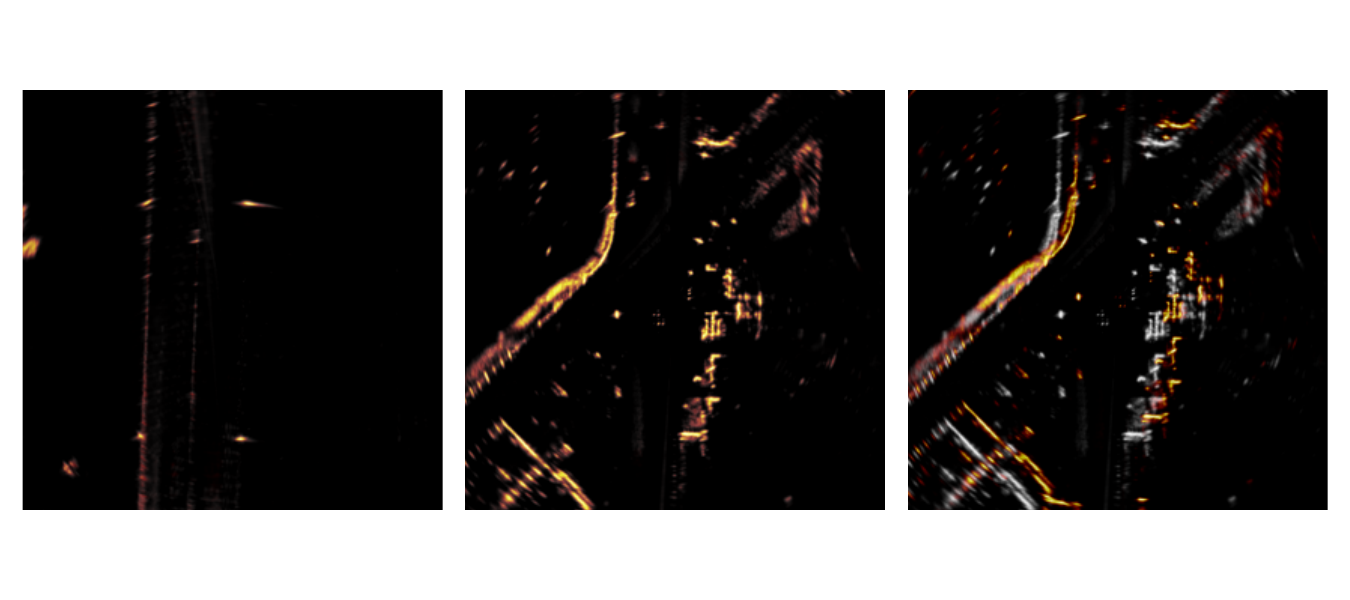}};
        \node[black, anchor=north west] at (-4.35,1.75) {\scriptsize \textbf{Scene A, good reg.}};
        \node[black, anchor=north west] at (-4.35,-1.30) {\scriptsize $g =\ $\SI{4.26e+08}{}};
        \node[Green, anchor=north west] at (-4.35,-1.6) {\scriptsize $s =\ $\SI{1.0}{}};
        
        \node[black, anchor=north west] at (-1.45,1.75) {\scriptsize \textbf{Scene B, good reg.}};        \node[black, anchor=north west] at (-1.45,-1.30) {\scriptsize $g =\ $\SI{1.89e+09}{}};
        \node[Green, anchor=north west] at (-1.45,-1.6) {\scriptsize $s =\ $\SI{0.98}{}};
        
        \node[black, anchor=north west] at (1.45,1.75) {\scriptsize \textbf{Scene B, poor reg.}};
        \node[black, anchor=north west] at (1.45,-1.30) {\scriptsize $g =\ $\SI{5.47e+08}{}};
        \node[Red, anchor=north west]   at (1.45,-1.6) {\scriptsize $s =\ $\SI{0.28}{}};
    \end{tikzpicture}
    %\vspace{-0.9cm}
    \caption{Example of local-map registration. Alone, the cross-correlation score $\corr$ does not allow for discrimination between good and poor registrations across all environments. The scaled score $s$ is a better proxy as it is less impacted by the scene's average intensity return.}
    \label{fig:correlation_scale}
\end{figure}

To refine the aforementioned coarse transformation, \ac{drpogo} performs direct registration via continuous cross-correlation maximization:
\begin{align}
    \begin{aligned}
    \transtilde{t_i}{t_j} &= \underset{\trans{t_i}{t_j}}{\text{argmax}}\quad \corr(\localmap_i, \localmap_j, \trans{t_i}{t_j}),\ \text{with}
    \\
    \corr(\localmap_i, \localmap_j, &\trans{t_i}{t_j}) = \sum_{x,y} \bilinear\left(\localmap_i, \trans{t_i}{t_j}\begin{bmatrix}
        x \\ y \\ 1
    \end{bmatrix}\right) \localmap_j(x,y),
    \label{eq:fine_reg}
    \end{aligned}
\end{align}
$x$ and $y$ some Cartesian coordinates in $\localmap_j$, $\localmap_j(x,y)$ the local-map intensity at coordinates $x,y$, and $\bilinear(\localmap_i, \begin{bmatrix}\hat{x} & \hat{y} & 1\end{bmatrix}^\top)$ the bilinear interpolation of $\localmap_i$'s intensity at coordinates $\hat{x},\hat{y}$.
While being a good relative metric for registration, the absolute cross-correlation value $\corr$ depends on the environment and the total `amount of intensity' in $\localmap_j$.
As illustrated in Fig.~\ref{fig:correlation_scale}, $\corr$ cannot be directly thresholded to classify a potential loop-closure registration as successful.
Thus, \ac{drpogo} applies a user-defined threshold over the scaled cross-correlation $s = \frac{\corr(\localmap_i, \localmap_j, \transtilde{t_i}{t_j})}{\sum_{x,y} \localmap_j^2(x,y)}$ to reject unlikely loop closures.
Intuitively, this score is a proxy for the overlap ratio between the non-zero-intensity data in $\localmap_i$ and $\localmap_j$.
In our experiments, the threshold is 0.5.

\subsection{Pose-graph optimization}

After coarse-to-fine registration of loop-closure candidates, the set of successful pairs $\loops$ and their associated relative transformations $\transtilde{t_i}{t_j}$ are integrated in an \ac{SE2} pose graph as binary factors with relative pose residuals:
\begin{equation}
    \resrel{i}{j} = \Log\left((\transtilde{t_i}{t_j})\inv(\trans{\world}{t_i})\inv\trans{\world}{t_j}\right),
\end{equation}
where $\Log$ converts an \ac{SE2} matrix into its $x$-$y$-$\theta$ vector representation.
The DRO scan-to-scan estimates are used to constrain consecutive poses with residuals $\resrel{m}{m+1}$.
If a gyroscope is used, DRO estimates angular velocity biases $\biastilde_m$ using a simple heuristic when the vehicle is stopped.
While improving over the scenario of no bias estimation, these bias estimates do not leverage radar information.
We propose to integrate bias corrections of DRO's biases in our pose-graph optimization by modifying DRO's scan-to-scan poses as $\transtilde{t_m}{t_{m+1}}\trans{b_m}{}$ with
\begin{equation}
    \trans{b_m}{} = \begin{bmatrix}
        \cos(\phi) & \sin(\phi) & 0 \\ -\sin(\phi) & \cos(\phi) & 0 \\ 0 & 0 & 1
    \end{bmatrix},\ \phi = (\bias_m - \biastilde_m)\Delta t.
    \nonumber
\end{equation}
In such a scenario, the state $\state$ to estimate is $\{\trans{\world}{t_1}, \bias_1,\cdots,\trans{\world}{t_N},\bias_N\}$.
Eventually, \ac{drpogo}'s pose-graph optimization is
\begin{equation}
    \statehat = \underset{\state}{\text{argmax}}\sum_{i=1}^{N-1}\left(\Vert\resrel{i}{i+1}\Vert^2_{\inforel{i}{i+1}} +\Vert\resbias{i}\Vert^2_{\infobias{i}}\right) + \sum_{\{i,j\}\in\loops}\Vert\resrel{i}{j}\Vert^2_{\inforel{i}{j}},
    \nonumber
\end{equation}
with $\Vert\mathbf{r}_\bullet\Vert_{\mathbf{\Omega}_\bullet} = \mathbf{r}_\bullet^\top\mathbf{\Omega}_\bullet\inv\mathbf{r}_\bullet$, $\mathbf{\Omega}_\bullet$ the covariance matrix associated with $\mathbf{r}_\bullet$, and the residual $\resbias{i}=\bias_i - \bias_{i+1}$ enforces the slow varying nature of gyroscope biases via a Brownian motion process.
To accommodate the presence of outliers in the loop-closure constraints, \ac{drpogo} applies a Cauchy loss function to the associated residuals.

\section{Experiments}

\subsection{Implementation details}

The original implementation of RaPlace computes the Radon transform for every incoming scan and attempts detection by computing a similarity score between the current scan and every previous frame.
However, the Radon transform is a slow operation ($\approx$ \SI{1}{\second}), and the $\mathcal{O}(N)$ matching strategy does not scale to very long trajectories.
To reduce the computational load, we only perform the Radon transform for keyframes and attempt loop-closure detection solely with past keyframes that are within a certain radius of the current estimate.
To maximize the number of loop-closure detections while allowing real-time operations, we dedicate one CPU core to RaPlace computations.
Accordingly, a local map is considered to be a new keyframe only if, upon reception, the RaPlace-dedicated CPU core has finished the previous detection.
For the match search, the radius is proportional to the distance travelled using the typical maximum drift of the odometry (arbitrarily set to \SI{1}{\percent}).

Our coarse, feature-based alignment is performed using OpenCV for feature extraction, matching (brute force), and \ac{SE2} registration.
The native registration method also estimates the scale alongside the \ac{SE2} transformation.
If the resulting scale differs from 1 ($\pm$ \SI{5}{\percent}), the pair of local maps is rejected.
The proposed fine registration extends DRO's GPU implementation using PyTorch.
Note that an additional coarse grid search around the estimate of the feature-based alignment step is performed to further avoid unwanted local minima.
Finally, the whole trajectory is estimated via pose-graph optimization every time a new loop-closure detection is validated by the direct registration refinement.
To accommodate long trajectory loops (which result in significant odometry drift), the optimization is performed once with a large scale for the Cauchy loss, then again with a smaller one.

\subsection{Datasets, baselines, and metrics}

\subsubsection{Datasets}

\begin{figure}
    \centering
    \def\legenddist{0.1cm}
    \def\vdist{0.6cm}
    \def\imgwidth{0.49\columnwidth}
    \def\hdist{0.02\columnwidth}
    \begin{tikzpicture}
        \tikzstyle{legend} = [align = center, inner sep=0, outer sep=0, node distance = 0em, execute at begin node=\setlength{\baselineskip}{8pt}\small] 
        \tikzstyle{image} = [align = center, inner sep=0, outer sep=0, node distance = 0em] 
        \node[image] (commercialA) {\includegraphics[clip, width=\imgwidth]{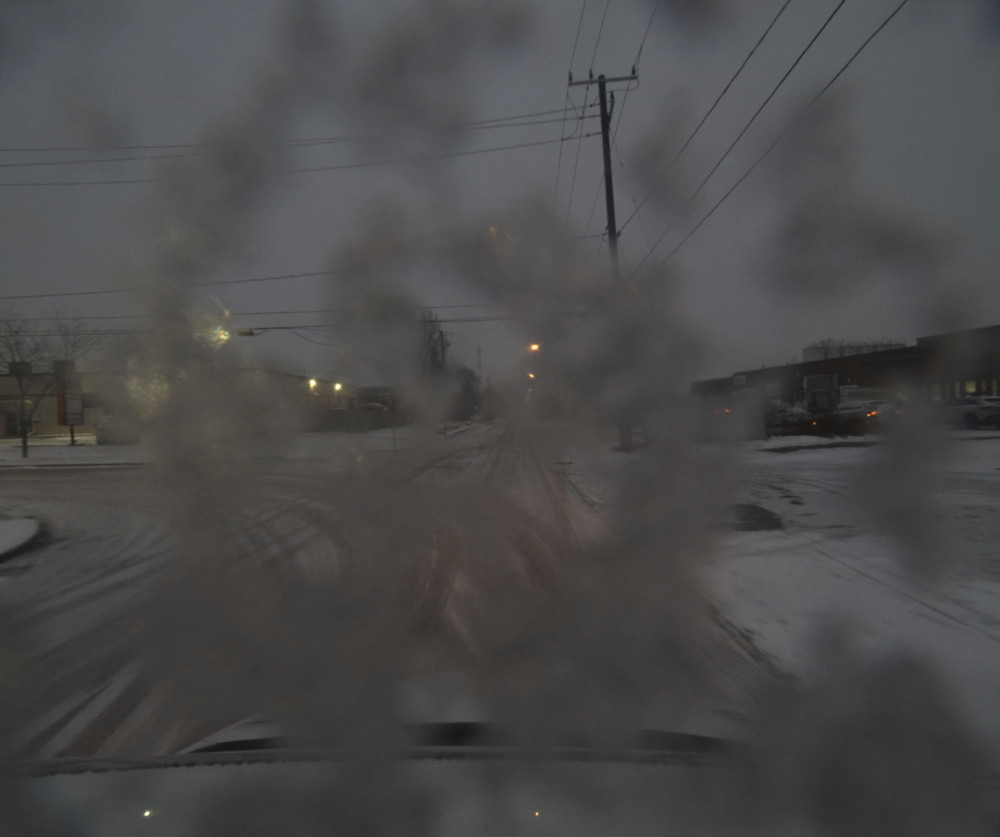}};
        \node[image, right=\hdist of commercialA] (commercialB) {\includegraphics[clip, width=\imgwidth]{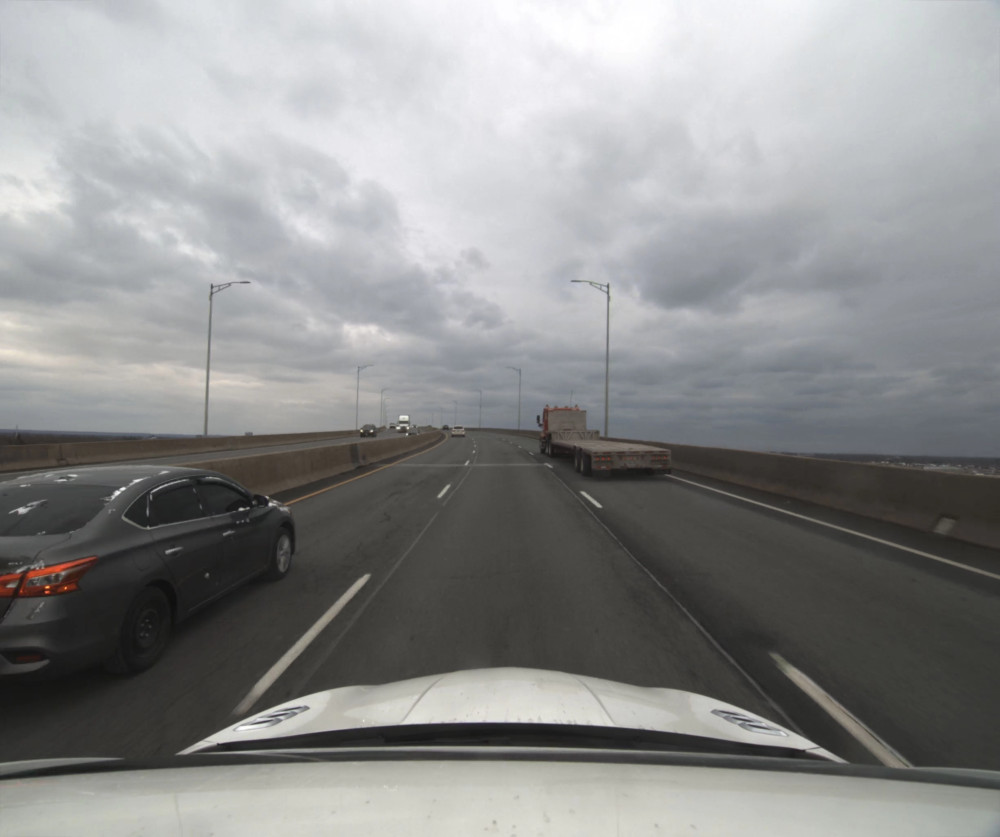}};
        
        \node[image, below=\vdist of commercialA] (skywayA) {\includegraphics[clip, width=\imgwidth]{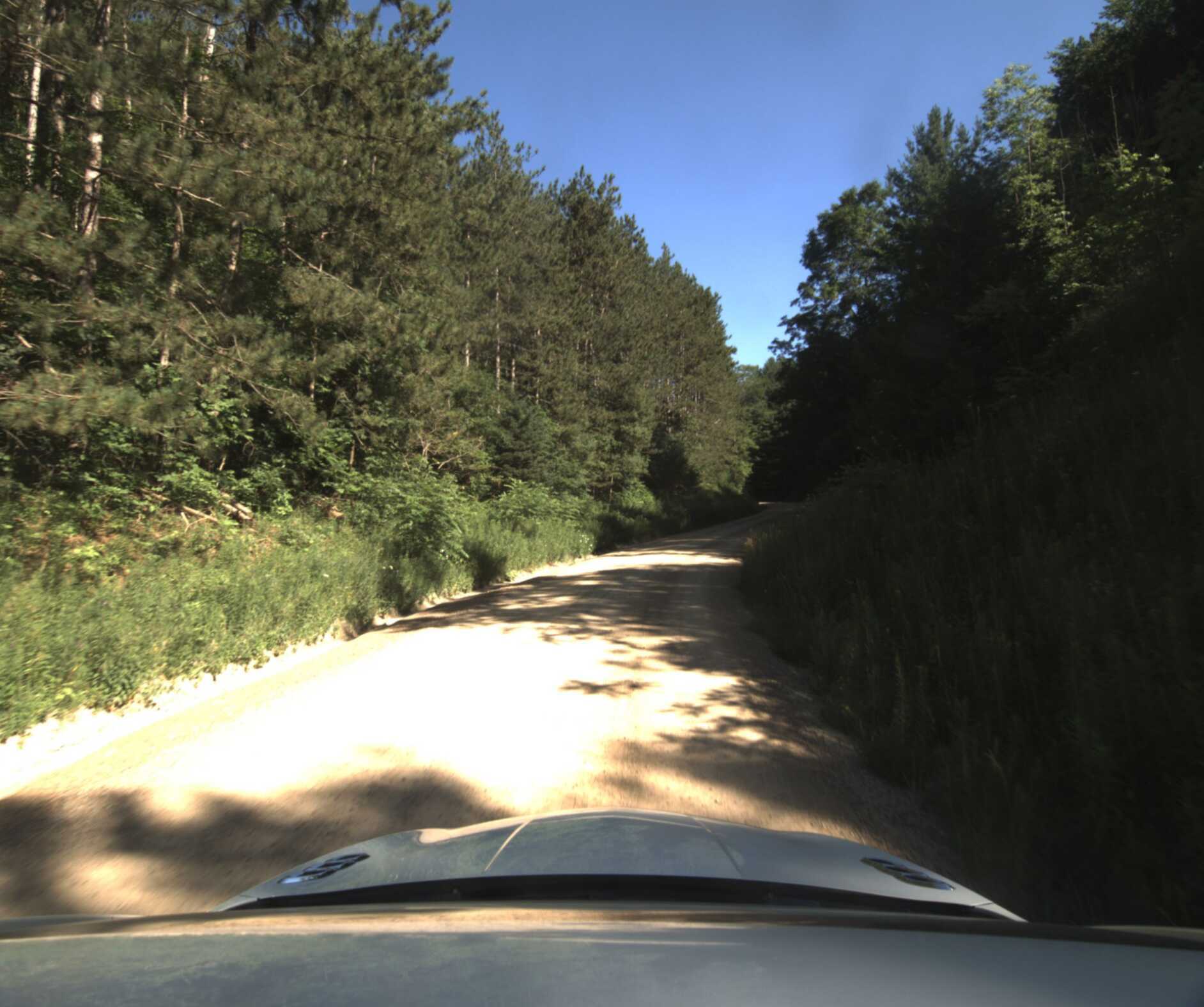}};
        \node[image, right=\hdist of skywayA] (skywayB) {\includegraphics[clip, width=\imgwidth]{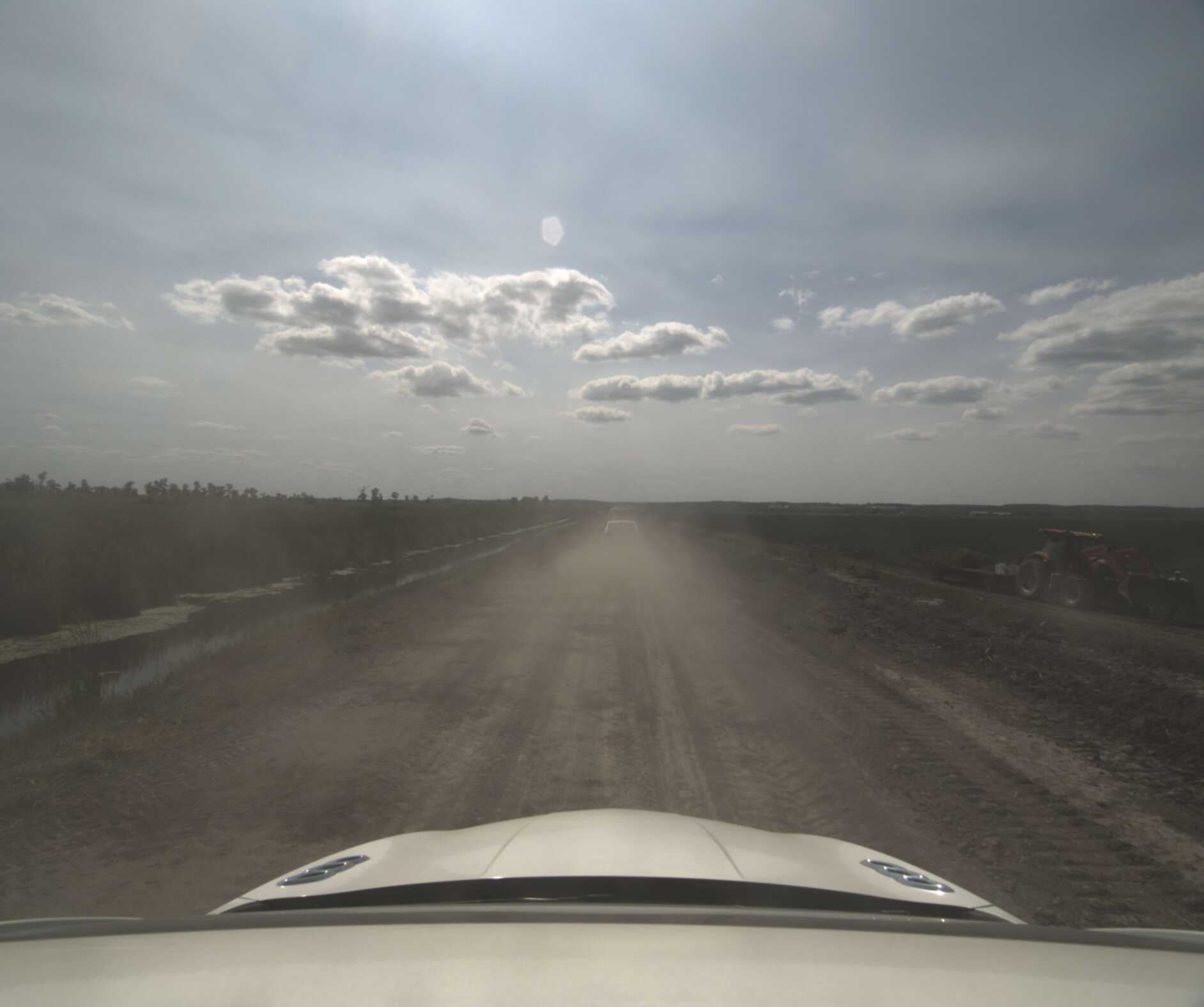}};

        \node[legend, below=\legenddist of commercialA] {(a) \texttt{Industrial} \scriptsize (with snowstorm)};
        \node[legend, below=\legenddist of commercialB] {(b) \texttt{Skyway} \scriptsize (weakly structured)};
        \node[legend, below=\legenddist of skywayA] {(c) \texttt{Forest} \scriptsize(natural environment)};
        \node[legend, below=\legenddist of skywayB] {(d) \texttt{Farm} \scriptsize(with dust)};
        
    \end{tikzpicture}
    \caption{Illustration of the challenging conditions in Boreas-RT}
    \label{fig:cam}
\end{figure}
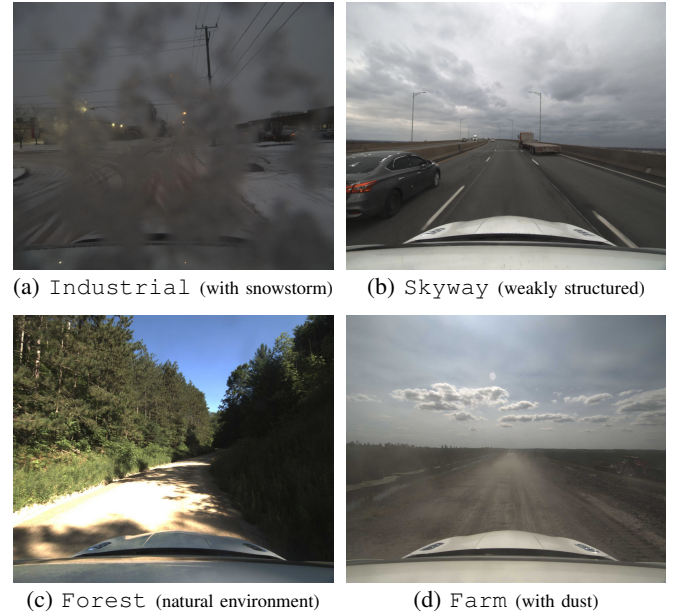

We have evaluated our method using both the Boreas~\cite{burnett2023boreas} and Boreas-RT~\cite{lisus2026boreasrt} datasets.
The former focuses on multi-season driving through a wide range of weather conditions on the same \SI{8}{\km} suburban route over a year, while the latter is collected along various routes with challenging roads and environments.
Both have been collected with an automotive platform equipped with a Navtech RAS6 radar and a gyroscope, among other sensors.
A notable difference is that the radar firmware used for Boreas-RT has a triangular modulation pattern that enables Doppler velocity extraction, as shown in~\cite{lisus2025doppler}.
For the evaluation with Boreas data, we randomly selected 13 sequences.\footnote{\texttt{2020-\{11-26-13-58}, \texttt{12-18-13-44\}}, \texttt{2021-\{01-26-11-22, 02-02-14-07, 03-02-13-38, 03-30-14-23, 04-20-14-11, 05-13-16-11, 07-20-17-33, 09-02-11-42, 10-15-12-35, 11-14-09-47, 11-23-14-27\}}.}
With Boreas-RT, we leverage 4 sequences of each of the following routes: \texttt{Suburbs} (\SI{7.9}{\kilo\meter}), \texttt{Industrial} (\SI{5.4}{\kilo\meter}), \texttt{Skyway} (\SI{11.1}{\kilo\meter}), \texttt{Forest} (\SI{16.4}{\kilo\meter}), and \texttt{Farm} (\SI{10.8}{\kilo\meter}).
Both the \texttt{Suburbs} and \texttt{Skyway} routes offer numerous opportunities for loop-closure detections as the second half of each sequence corresponds to the same itinerary as the first half, but in the other direction.
The other sequences form large loops with a fairly limited overlap between the beginning and end of the trajectories.
As illustrated in Fig.~\ref{fig:cam}, the selected environments span a wide range of challenging driving conditions with snowstorms, weakly structured environments, and natural scenes.
Additionally, the \texttt{Forest} and \texttt{Farm} sequences are collected on uneven dirt roads, inducing significant roll and pitch of the vehicle.

\subsubsection{Baselines}

We benchmark \ac{drpogo} against two recent open-source radar \ac{slam} implementations.
The first is Navtech-SLAM~\cite{kim2024navtechslam}, which relies on ORORA~\cite{lim2023orora} for odometry, Scan Context~\cite{kim2018scancontext} for loop-closure detection and ICP for loop-closure registration.
The second is TBV-SLAM, with CFEAR-3~\cite{adolfsson2023cfear} as the odometry method and Scan Context~\cite{kim2018scancontext} with CorAl~\cite{adolfsson2022coral} for loop-closure detection/validation.
Note that these two methods also rely on a pose-graph optimization for full-batch trajectory estimation.

\subsubsection{Metrics}

The \ac{ate} is used to evaluate the overall accuracy of \ac{drpogo} and the different baselines for trajectory estimation.
After aligning the ground-truth trajectory and the estimated one with SVD-based \ac{SE2} registration, the \ac{ate} is computed as $\sqrt{\frac{1}{N}\sum_{i=1}^{N}\Vert\poshat{\world}{t_i} - \pos{\gt}{t_i}\Vert^2}$, with $\poshat{\blacksquare}{\bullet}$ the translation component of $\transhat{\blacksquare}{\bullet}$.
Since all the sequences finish close to the starting location, we use the \ac{epe} as a proxy to assess the effectiveness of loop-closure constraints in each method.
The \ac{epe} is defined as the norm of the translation component of $(\trans{\gt}{t_N})\inv\trans{\gt}{t_1}(\transhat{\world}{t_1})\inv\transhat{\world}{t_N}$.

\subsection{Trajectory error}
\subsubsection{Boreas dataset}

\begin{figure}
    \centering
    \includegraphics[width=\columnwidth]{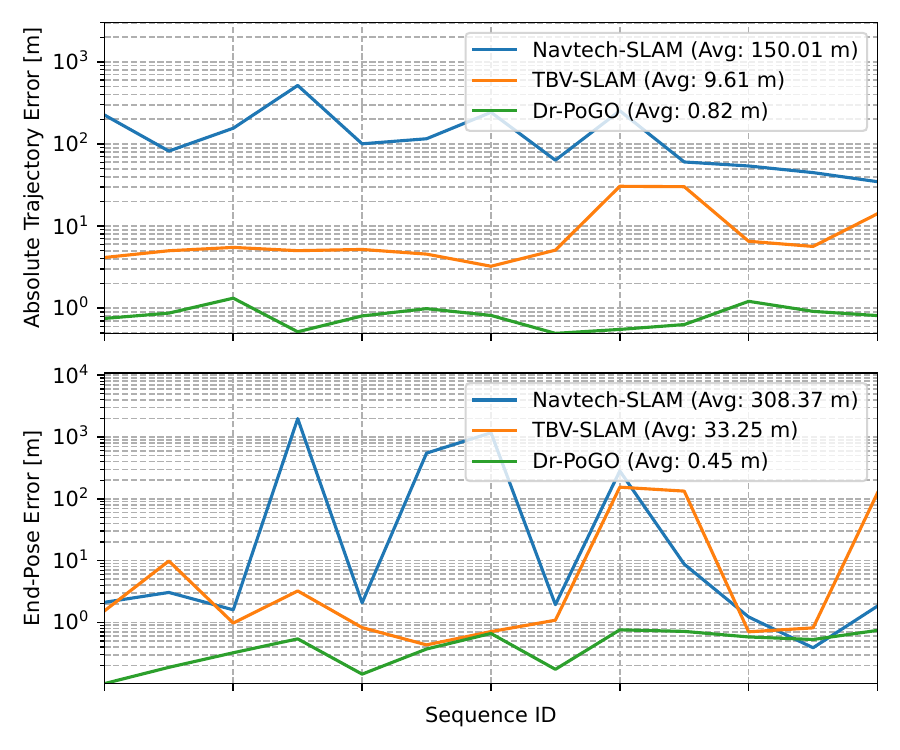}
    \caption{Absolute Trajectory Error and End-Pose Error (log scale) for each of the 13 sequences used for benchmarking Dr-PoGO and the baselines on the Boreas dataset.}
    \label{fig:publicboreas}
\end{figure}

Fig.~\ref{fig:publicboreas} shows the per-sequence \ac{ate} and \ac{epe} for Dr-PoGO, Navtech-SLAM and TBV-SLAM on the Boreas dataset~\cite{burnett2023boreas}.
\ac{drpogo} achieves the lowest \ac{ate}, with an average of \SI{0.82}{\meter} and a maximum of \SI{1.32}{\meter}.
In terms of \ac{epe}, \ac{drpogo} also outperforms the different baselines.
TBV-SLAM attains moderate \acp{ate} on average (\SI{9.61}{\meter}) but most sequences are between \SI{2}{} and \SI{3}{\meter} \ac{ate}.
There is a strong correlation between the \ac{ate} and \ac{epe} in TBV-SLAM's results, suggesting that the odometry (CFEAR~\cite{adolfsson2023cfear}) is fairly reliable, but occasional failures in loop-closure detection hinder the system's global performance.
The presence of traffic in the data and the conservative nature of CorAl~\cite{adolfsson2022coral} loop-closure validation are possible explanations for missing loop closures.
The inconsistency in Navtech-SLAM's results and the absence of a clear correlation between \ac{ate} and \ac{epe} indicate a lack of robustness both in the odometry (ORORA~\cite{lim2023orora}) and the loop-closure detection mechanism.
Indeed, ORORA relies on a feature-matching front end that is sensitive to sparse or degraded radar scans, as it reduces both the number and the spatial distribution of correspondences.
This negatively impacts the decoupled rotation estimates and subsequent anisotropic translations.

Overall, we attribute \ac{drpogo}'s performance to a more reliable odometry source (DRO~\cite{legentil2025dro}) and our coarse-to-fine loop-closure registration and validation using local maps.
The use of local maps improves robustness in real-world scenarios where dynamic elements, such as traffic, impede scan-based operations.
Additionally, using the local-map scaled cross-correlation to validate or not a loop-closure candidate provides more information than a descriptor-based similarity score.
Quantitative results of an ablation study in Section~\ref{sec:ablation} support these claims.

\subsubsection{Boreas-RT dataset}

Unfortunately, when using radar data collected with a triangular modulation pattern, the odometry component of Navtech-SLAM fails.
To obtain quantitative results for Navtech-SLAM, we preprocessed the radar data by removing the Doppler distortion based on the ground-truth velocities.
Table~\ref{tab:selfcollected} presents both the average \ac{ate} and \ac{epe} for each method in each environment.
The figures in this table only account for successful runs.
Defining success by an \ac{ate} under 1\% of the travelled distance, Navtech-SLAM and TBV-SLAM fail in some sequences.
Omitting these, all the \ac{slam} methods correct large odometry drifts thanks to successful loop-closure detections in \texttt{Suburbs} and \texttt{Industrial} environments, as shown by the low \ac{epe}.
As expected, the average \ac{epe} obtained with DRO is large due to the absence of a loop-closure mechanism.
Thanks to DRO's robustness, \ac{drpogo} is the only \ac{slam} method to be successful on the highly challenging \texttt{Skyway}, \texttt{Forest}, and \texttt{Farm} sequences.

\begin{figure}
    \centering
    \includegraphics[width=\columnwidth]{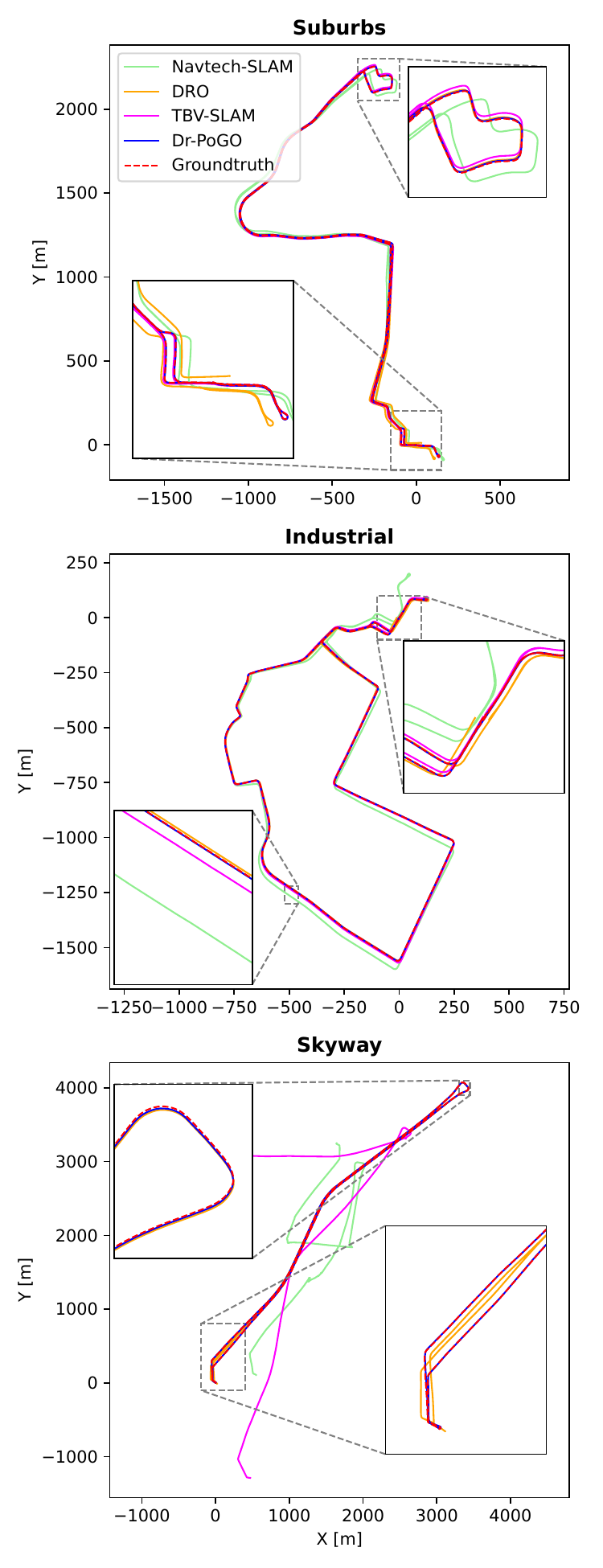}
    \caption{Trajectory examples over the three environment types of the self-collected dataset. Estimated trajectories are aligned to the ground-truth as for the ATE computation.}
    \label{fig:trajectories}
\end{figure}

When considering the \ac{ate}, the different approaches span a wider range of results.
Navtech-SLAM performs the worst.
As illustrated in Fig.~\ref{fig:trajectories}, it suggests that ORORA odometry introduces a drift/warp of the trajectory that is too large to be fully corrected with loop-closure constraints.
TBV-SLAM performs well with the second-best results.
The estimated trajectories are more consistent with the true motion when compared to Navtech-SLAM.
\ac{drpogo} significantly outperforms all the other methods with GPS-level ATE ($<$ \SI{5}{\meter}) throughout all setups.
Comparing \ac{drpogo}'s results (\ac{ate} and \ac{epe}) with DRO's demonstrates the soundness of our approach and its ability to reach new levels of robust and accurate trajectory estimation.

The trajectory accuracy displayed by \ac{drpogo} rivals that of lidar-based frameworks.
To support that claim, we have selected Fast-LIO2~\cite{xu2022fastlio2} and 2Fast-2Lamaa~\cite{legentil20242fast2lamaa} as lidar-inertial baselines.
These two frameworks do not perform explicit loop closure but estimate their state in a scan-to-map manner.
Accordingly, for trajectories where the vehicle retraces its way to the starting location, it can be seen as loop-closure localization for every lidar scan.
Unfortunately, when the trajectory forms a large loop, these methods fail due to odometry drift.
Thus, we only benchmark Fast-LIO2 and 2Fast-2Lamaa on the \texttt{Suburbs} sequences and obtain \acp{ate}\footnote{
To convert the SE(3) estimates into SE(2), we first fit a plane to the estimated trajectories and project all the poses on that plane.} of \SI{13.82}{}\footnote{Note that Fast-LIO2's results are computed on 3 out of the 5 sequences due to dramatic failures, probably caused by unhandled dynamic objects.} and \SI{2.26}{\meter}, respectively.
\ac{drpogo} offers a four-fold improvement over the best lidar baseline.

\begin{table*}
    \centering
    \caption{Average ATE and EPE [m] on self-collected dataset}
    \setlength{\tabcolsep}{2pt}
    \small
    \begin{tabularx}{\linewidth}{lYYYYY}
        \toprule
        \textbf{Method} & \textbf{Suburbs} & \textbf{Industrial} & \textbf{Skyway} & \textbf{Forest} & \textbf{Farm}
        \\
        \midrule
        DRO \cite{legentil2025dro} & 9.53 / 38.9 & 6.32 / 17.2 & 18.6 / 60.7 & 25.8 / 80.5 & 38.8 / 170.4
        \\
        Navtech-SLAM~\cite{kim2024navtechslam}$^*$ & 36.4 / 0.65$^{1}$ & 28.9 / 1.46$^1$ & - / -$^4$ & - / -$^4$ & - / -$^4$
        \\
        TBV-SLAM \cite{adolfsson2023tbv} & 6.11 / 0.55 & 4.72 / 0.82$^1$ & - / -$^4$ & - / -$^4$ & - / -$^4$
        \\
        Dr-PoGO (ours) & \textbf{0.75} / \textbf{0.42} & \textbf{1.58} / \textbf{0.77} & \textbf{3.16} / \textbf{0.23} & \textbf{4.31} / \textbf{0.75} & \textbf{4.19} / \textbf{0.56}
        \\
        \bottomrule
        \multicolumn{6}{c}{\scriptsize Results reported as XX / YY, with XX the ATE and YY the EPE.}
        \\
        \multicolumn{6}{c}{\scriptsize $^1$ The superscript indicates the number of failed sequences (ATE above 1\% of the sequence distance). \quad $^*$ Navtech-SLAM required Doppler undistortion preprocessing.}
    \end{tabularx}
    \label{tab:selfcollected}
\end{table*}

\subsection{Loop-closure detection and registration}
\label{sec:exploopclosure}

\begin{figure}
    \centering
    \includegraphics[clip, width=\columnwidth, trim=0.3cm 0.3cm 0.3cm 0.3cm]{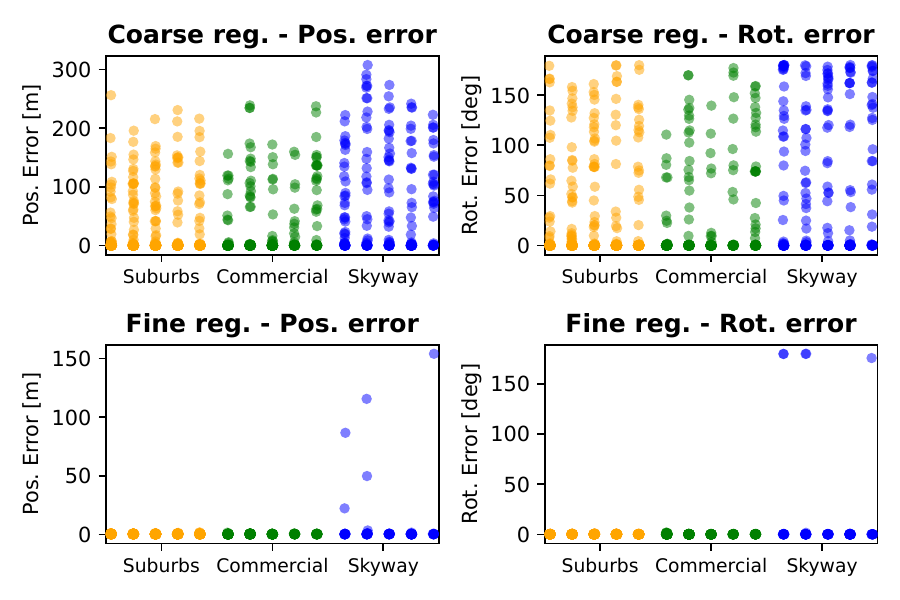}
    \caption{Strip plots of the position (left column) and rotation (right column) alignment error after the coarse feature-based registration (top row) and the direct refinement (bottom row) using the 15 self-collected sequences.}
    \label{fig:loopclosure}
\end{figure}

In this section, we analyze the accuracy of the proposed coarse-to-fine loop-closure registration on the three most structured sequences of the Boreas-RT dataset.
Fig.~\ref{fig:loopclosure} shows strip plots of the alignment errors for the 15 self-collected sequences, and Table~\ref{tab:loopclosure} provides numerical values per environment type: average per-sequence number of matches, average number of inliers (position and rotation errors under \SI{1.5}{\meter} and \SI{1}{\degree}, respectively), position and rotation \ac{rmse} for the inliers.
The \ac{rmse} values suggest that direct refinement significantly improves registration accuracy for all sequence types compared to feature-based alignment.
One can see that after the coarse registration, there is a large number of outliers.
It is important to remember that \ac{drpogo} does not threshold RaPlace's similarity score to keep a maximum number of true-positive matches and prevent having too many parameters to tune.
Thus, it is normal to find many outliers after the coarse registration step.
When applying the direct pose refinement with our proposed scaled-cross-correlation score threshold, the number of outliers is drastically reduced with a very limited impact on the number of inliers, thus greatly improving the inlier ratio.

\begin{table}
    \centering
    \caption{Loop-closure registration analysis}
    \setlength{\tabcolsep}{2pt}
    \small
    \begin{tabularx}{\linewidth}{lYYYYY}
        \toprule
        \textbf{Seq. type} & \textbf{Avg. matches} & \textbf{Avg. inliers} & \textbf{Inlier ratio} & \textbf{In. pos. RMSE} & \textbf{In. rot. RMSE}
        \\
        \midrule
        \multicolumn{6}{c}{\textit{Coarse registration}} 
        \\
        \midrule
        Suburbs & 130 & \textbf{101} & 78\% & 0.50$\,\mathrm{m}$ & 0.26$^\circ$
        \\
        Industrial & 72.6 & \textbf{54.8} & 75\% & 0.38$\,\mathrm{m}$ & 0.26$^\circ$
        \\
        Skyway & 43.2 & 17.0 & 39\% & 0.51$\,\mathrm{m}$ & 0.26$^\circ$
        \\
        \midrule
        \multicolumn{6}{c}{\textit{Fine registration}} 
        \\
        \midrule
        Suburbs & 91.0 & 90.6 & \textbf{99}\% & \textbf{0.34}$\,\mathrm{m}$ & \textbf{0.21}$^\circ$
        \\
        Industrial & 55.4 & 54.2 & \textbf{98}\% & \textbf{0.27}$\,\mathrm{m}$ & \textbf{0.19}$^\circ$
        \\
        Skyway & 19.4 & \textbf{18.2} & \textbf{94}\% & \textbf{0.36}$\,\mathrm{m}$ & \textbf{0.17}$^\circ$
        \\
        \bottomrule
        \multicolumn{6}{c}{\scriptsize The best results between coarse and fine registration are shown in bold.}
    \end{tabularx}
    \label{tab:loopclosure}
\end{table}

\subsection{Ablation study}
\label{sec:ablation}

To provide insight into the impact of each component of the proposed method, we have conducted an ablation study using the \texttt{Suburbs}, \texttt{Industrial}, and \texttt{Skyway} sequences of the Boreas-RT dataset.
Table~\ref{tab:ablation} shows the corresponding \acp{ate} and \acp{epe}.
Note that all variants share the same parameters.
The first observation is the importance of using a gyroscope; without it, \ac{drpogo}'s performance significantly drops both in terms of \ac{ate} and \ac{epe}, and some failures occur.
It suggests that the accuracy of gyro-less odometry is not sufficient to provide high-quality trajectory priors and, thus, impedes loop-closure detections, leading to missing or erroneous loop-closure constraints.
Further analysis of gyro-less operations is part of future work. 
When using a gyroscope, the use of local maps (both for odometry and loop-closure detection/registration) and the direct refinement are \ac{drpogo}'s most critical features.
Not leveraging local maps or relying solely on coarse loop-closure registration results in errors up to four times higher.
Finally, thanks to the relatively high quality of the gyroscope present on our data collection platform, the impact of gyroscope bias is fairly limited.
Accordingly, ignoring the bias does not have a significant impact on \ac{drpogo}'s overall performance.

\begin{table}
    \centering
    \caption{Dr-PoGO ablation study (average ATE and EPE [m])}
    \setlength{\tabcolsep}{2pt}
    \small
    \begin{tabularx}{\linewidth}{lYYY}
        \toprule
        \textbf{Variant} & \textbf{Suburbs} & \textbf{Industrial} & \textbf{Skyway}
        \\
        \midrule
        No local map & 2.97 / 10.38 & 6.61 / 19.5 & 3.8 / 0.35
        \\
        Coarse reg. & 1.53 / 0.72$^1$ & 2.16 / 1.63 & 3.96 / 0.18$^1$
        \\
        No bias & 0.76 / 0.48 & 1.62 / 0.90 & \textbf{3.15} / \textbf{0.21}
        \\
        No gyro & 7.05 / 4.09$^1$ & 9.21 / 1.59$^2$ & - / -$^4$
        \\
        Dr-PoGO (Ours) & \textbf{0.75} / \textbf{0.42} & \textbf{1.58} / \textbf{0.77} & 3.16 / 0.23
        \\
        \bottomrule
        \multicolumn{4}{c}{\scriptsize Results reported as XX / YY, with XX the ATE and YY the EPE.}
        \\
        \multicolumn{4}{c}{\scriptsize $^1$ The superscript indicates the number of failed sequences.}
    \end{tabularx}
    \label{tab:ablation}
\end{table}

\subsection{Computation time}

Table~\ref{tab:computation} shows the breakdown of \ac{drpogo}'s computation time using a laptop equipped with an Intel i7-13850HX CPU and an Nvidia RTX 5000 Mobile GPU.
The column `time per operation' represents the marginal cost of each component.
In contrast, the last column presents the average time per frame, calculated by dividing the total time by the number of radar scans in each sequence.
Using a \SI{4}{\hertz} radar as in the Boreas and self-collected datasets, we demonstrated that \ac{drpogo} runs in real-time, as the sums of GPU and CPU times per frame are below the \SI{250}{\milli\second} mark. 
Note that our implementation of RaPlace loop-closure detection attempts to maximize the usage of 1 CPU core.
For lightweight operations, one can reduce the number of keyframes for place recognition and subsequent registrations and optimizations.

\begin{table}
    \centering
    \caption{Dr-PoGO computation time analysis}
    \setlength{\tabcolsep}{2pt}
    \small
    \begin{tabularx}{\linewidth}{lYYY}
        \toprule
        \textbf{Component} & \textbf{Hardware} & \textbf{Time per operation [s]} & \textbf{Avg. per raw frame [s]}
        \\
        \midrule
        DRO-GD & GPU & 0.104 & 0.104
        \\
        RaPlace & CPU (1 core) & 0.646 & 0.215
        \\
        Coarse reg. & CPU (1 core) & 0.170 & 0.024
        \\
        Direct reg. & GPU & 0.389 & 0.013
        \\
        Pose graph & CPU (1 core) & 0.093 & 0.002
        \\
        \bottomrule
    \end{tabularx}
    \label{tab:computation}
\end{table}

\section{Conclusion}

This paper introduced \ac{drpogo}, a framework for radar-based 2D \ac{slam}.
It relies on DRO for odometry, RaPlace for loop-closure detection, and introduces a novel coarse-to-fine local-map alignment pipeline that leverages feature-based and direct registration techniques.
The \ac{SE2} trajectory is estimated through a full-batch pose-graph optimization.
To the best of our knowledge, it is the first radar \ac{slam} approach that leverages direct registration for odometry and loop-closure registration. 
We have benchmarked our framework over \SI{300}{\kilo\meter} of real-world automotive data (sequences up to \SI{16}{\kilo\meter}-long) and demonstrated state-of-the-art trajectory accuracy.
We even showed that \ac{drpogo} can outperform state-of-the-art lidar-based frameworks.
To enable direct localization with respect to a single global map, future work includes using \ac{drpogo}'s output as the initial guess for a global bundle adjustment optimization that directly leverages radar data in a single optimization instead of \ac{SE2} relative transformations.

\section{Acknowledgement}

The authors would like to thank Daniil Lisus and Katya M. Papais for leading the Boreas-RT collection campaign, and Raphael Falque for running the lidar baselines.

\bibliographystyle{IEEEtran}
\bibliography{references}


%%%%%%%%%%%%%%%%%%%%%%%%%%%%%%%%%%%%%%%%%%%%%%%%%%%%%%%%%%%%%%%%%%%%%%%%%%%%%%%%

%%%%%%%%%%%%%%%%%%%%%%%%%%%%%%%%%%%%%%%%%%%%%%%%%%%%%%%%%%%%%%%%%%%%%%%%%%%%%%%%

%%%%%%%%%%%%%%%%%%%%%%%%%%%%%%%%%%%%%%%%%%%%%%%%%%%%%%%%%%%%%%%%%%%%%%%%%%%%%%%%

\end{document}